# FUZZY ARTMAP AND NEURAL NETWORK APPROACH TO ONLINE PROCESSING OF INPUTS WITH MISSING VALUES

F.V. Nelwamondo* and T. Marwala*

*School of Electrical and Information Engineering, University of the Witwatersrand, Johannesburg, Private Bag 3, Wits, 2050, South Africa.

**Abstract:** An ensemble based approach for dealing with missing data, without predicting or imputing the missing values is proposed. This technique is suitable for online operations of neural networks and as a result, is used for online condition monitoring. The proposed technique is tested in both classification and regression problems. An ensemble of Fuzzy-ARTMAPs is used for classification whereas an ensemble of multi-layer perceptrons is used for the regression problem. Results obtained using this ensemble-based technique are compared to those obtained using a combination of auto-associative neural networks and genetic algorithms and findings show that this method can perform up to 9% better in regression problems. Another advantage of the proposed technique is that it eliminates the need for finding the best estimate of the data, and hence, saves time.

**Key words:** Autoencoder neural networks, Fuzzy-ARTMAP, Genetic algorithms, Missing data, MLP

## 1. INTRODUCTION

Real time processing applications that are highly dependent on the newly arriving data often suffer from the problem of missing data. In cases where decisions have to be made using computational intelligence techniques, missing data become a hindering factor. The biggest challenge on one hand is that most computational intelligence techniques such as neural networks are not able to process input data with missing values and hence, cannot perform classification or regression when some input data are missing. Various heuristics for missing data have however been proposed in the literature [1]. The simplest method is known as 'listwise deletion' and this method simply deletes instances with missing values [1]. The major disadvantage of this method is the dramatic loss of information in data sets. There is also a well documented evidence showing that ignorance and deletion of cases with missing entries is not an effective strategy [1-2]. Other common techniques are imputation methods based on statistical procedures such as mean computation, imputing the most dominant variable in the database, hot deck imputation and many more. Some of the best imputation techniques include the Expectation Maximization (EM) algorithm [3] as well as neural networks coupled with optimisation algorithms such as genetic algorithms as used in [4] and [5]. Imputation techniques where missing data are replaced by estimates are increasingly becoming popular. A great deal of research has been done to find more accurate ways of approximating these estimates. Among others, Abdella and Marwala [4] used neural networks together with Genetic Algorithms (GA) to approximate missing data. Gabrys [6] has also used Neuro-fuzzy techniques in the presence of missing data for pattern recognition problems.

The other challenge in this work is that, online condition monitoring uses time series data and there is often a limited time between the readings depending on how frequently the sensor is sampled. In classification and regression tasks, all decisions concerning how to proceed must be taken during this finite time period. Methods using optimisation techniques may take longer periods to converge to a reliable estimate and this depends entirely on the complexity of the objective function being optimised. This calls for better techniques to deal with this missing data problem.

We argue in this paper that it is not always necessary to have the actual missing data predicted. Differently said, it is not in all cases that the decision is dependent on *all* actual values. Therefore, a vast amount of computational resources is wasted in attempts to predict the missing values, whereas the ultimate result could have been achieved without such values. In light of this challenge, this paper investigates a problem of condition monitoring where computational intelligence techniques are used to classify and regress in the presence of missing data without the actual prediction of missing values. A novel approach where no attempt is made to recover the missing values, for both regression and classification problems, is presented. An ensemble of fuzzy-ARTMAP classifiers to classify in the presence of missing data is proposed. The algorithm is further extended to a regression application where Multi-layer Perceptron (MLP) is used in an attempt to get the correct output with limited input variables. The proposed method is compared to a technique that combines neural networks with Genetic Algorithm (GA) to approximate the missing data.

## 2. MISSING DATA THEORY

According to Little and Rubin [1], missing data are categorized into three basic types namely: 'Missing at Random', (MAR), 'Missing Completely at Random', (MCAR) and 'Missing Not at Random', (MNAR). MAR is also known as the ignorable case [3]. The probability of datum *d* from a sensor *S* to be missing at random is dependent on other measured variables from other sensors. A simple example of MAR is when sensor *T* is only read if sensor *S* reading is above a certain threshold. In this case, if the value read from sensor *S* is below the threshold, there will be no need to read sensor *T* and hence, readings from *T* will be declared missing at random. MCAR on the other hand refers to a condition where the probability of *S* values missing is independent of any observed data. In this regard, the missing value is neither dependent on the previous state of the sensor nor any reading from any other sensor. Lastly, MNAR occurs when data is neither MAR nor MCAR and is also referred to as the non-ignorable case [1, 3] as the missing observation is dependent on the outcome of interest. A detailed description of missing data theory can be found in [3]. In this paper, we shall assume that data is MAR.

## 3. BACKGROUND

*3.1 Neural network: multi-layer perceptrons*

Neural networks may be viewed as systems that learn the complex input-output relationship from any given data. The training process of neural networks involves presenting the network with inputs and corresponding outputs and this process is termed *supervised learning*. There are various types of neural networks but we shall only discuss the MLP since they are used in this study. MLPs are feed-forward neural networks with an architecture comprising of the input layer, hidden layer and the output layer. Each layer is formed from smaller units known as neurons. Neurons receive the input signals *x* and propagate them forward to the network and maps the complex relationship between inputs and the output. The first step in approximating the weight parameters of the model is finding the approximate architecture of the MLP, where the architecture is characterized by the number of hidden units, the type of activation function, as well as the number of input and output variables. The second step estimates the weight parameters using the training set [7]. Training estimates the weight vector $\vec{W}$ that ensures that the output is as close to the target vector as possible. This paper implements the autoencoder neural network as discussed below.

*Autoencoder neural networks*: Autoencoders, also known as auto-associative neural networks, are neural networks trained to recall the input space. Thompson *et al* [8] distinguish two primary features of an autoencoder network, namely the auto-associative nature of the network and the presence of a bottleneck that occurs in the hidden layers of the network, resulting into a butterfly-like structure. In cases where it is necessary to recall the input, autoencoders are preferred due to their remarkable ability to learn certain linear and non-linear interrelationships such as correlation and covariance inherent in the input space. Autoencoders project the input onto some smaller set by *intensively squashing* it into smaller details. The optimal number of the hidden nodes of the autoencoder, though dependent on the type of application, must be smaller than that of the input layer [8]. Autoencoders have been used in various applications including the treatment of missing data problem by a number of researchers including [4] and [9].

In this paper, auto-encoders are constructed using the MLP networks and trained using back-propagation. The structure of an autoencoder constructed using an MLP network is shown in Figure 1. The first step in approximating the weight parameters of the model is finding the approximate architecture of the MLP, where the architecture is characterized by the number of hidden units, the type of activation function, as well as the number of input and output variables. The second step estimates the weight parameters using the training set [7].

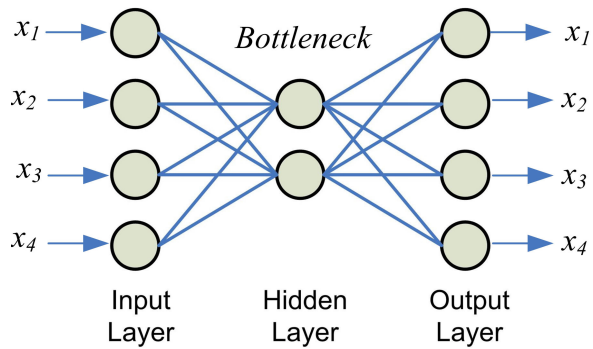

Figure 1: The structure of a four-input four-output autoencoder

Training estimates the weight vector $\vec{W}$ to ensure that the output is as close to the target vector as possible. The problem of identifying the weights in the hidden layers is solved by maximizing the probability of the weight parameter using Bayes' rule [8] as follows:

$$p(\vec{W} \mid D) = \frac{P(D \mid \vec{W})P(\vec{W})}{P(D)} \quad (1)$$

Where:

D is the training data, $P(\vec{W}|D)$ is the posterior probability, $P(D|\vec{W})$ is called the likelihood term that balances between fitting the data well and helps in avoiding overly complex models whereas $P(\vec{W})$ is the prior probability of $\vec{W}$ and $P(D)$ is the evidence term that normalizes the posterior probability. The input is transformed from *x* to the middle layer, *a*, using weights $w_{ij}$ and biases $b_i$ as follows [8]:

$$a_j = \sum_{i=1}^{d} \vec{W}_{ji} x_i + b_j \qquad (2)$$

where *j* = 1 and *j* = 2 represent the first and second layer respectively. The input is further transformed using the activation function such as the hyperbolic tangent (*tanh*) or the sigmoid in the hidden layer. More information on neural networks can be found in [10].

*3.2 Genetic Algorithms*

Genetic algorithms use the concept of survival of the fittest over consecutive generations to solve optimisation problems [11]. As in biological evolution, the fitness of each population member in a generation is evaluated to determine whether it will be used in the breeding of the next generation. In creating the next generation, the use of techniques (such as inheritance, mutation, natural selection, and recombination) common in the field of evolutionary biology are employed. The GA algorithm implemented in this paper uses a population of string chromosomes, which represent a point in the search space [11]. In this paper, all GA parameters were empirically determined. GA is implemented by following three main procedures which are selection, crossover and mutation. The algorithm listing in Figure 2 illustrates how GA operates.

| GA Algorithm |
| --- |
| 1). *Create an initial population $P$, beginning at an initial generation $g = 0$.* |
| 2). *for each population P, evaluate each population member (chromosome) using the defined fitness evaluation function possessing the knowledge of the competition environment.* |
| 3). *using genetic operators such as inheritance, mutation and crossover, alter $P(g)$ to produce $P(g+1)$ from the fit chromosomes in P (g).* |
| 4). *repeat steps (2) and (3) for the number of generations $G$ required.* |

Figure 2: Schematic representation of the Genetic algorithm operation

*3.3 Fuzzy ARTMAP*

Fuzzy ARTMAP is a neural network architecture developed by Carpernter *et al* [12] and is based on Adaptive Resonance Theory (ART). The Fuzzy ARTMAP has been used in condition monitoring by Javadpour and Knapp [13], but their application was not online. The Fuzzy ARTMAP architecture is capable of fast, online, supervised incremental learning, classification and prediction [12]. The fuzzy ARTMAP operates by dividing the input space into a number of hyperboxes, which are mapped to an output space. Instance based learning is used, where each individual input is mapped to a class label. Three parameters namely the vigilance $\rho \in [0, 1]$, the learning rate $\beta \in [0, 1]$ and the choice parameter $\alpha$, are used to control the learning process. The choice parameter is generally made small and a value of 0.01 was used in this application. The parameter $\beta$ controls the adaptation speed, where 0 implies a slow speed and 1, the fastest. If $\beta = 1$, the hyperboxes get enlarged to include the point represented by the input vector. The vigilance represents the degree of belonging and it controls how large any hyperbox can become, resulting in new hyperboxes being formed. Larger values of $\rho$ lead to a case where smaller hyperboxes are formed and this eventually lead to 'category proliferation', which can be viewed as overtraining. A complete description of the Fuzzy ARTMAP is provided in [12]. In this work, Fuzzy ARTMAP is preferred due to its incremental learning ability. As new data is sampled, there will be no need to retrain the network as would be the case with the MLP.

### 4. NEURAL NETWORKS AND GENETIC ALGORITHM FOR MISSING DATA

The method used here combines the use of auto-associative neural networks with genetic algorithms to approximate missing data. This method has been used by Abdella and Marwala [4] to approximate missing data in a database. A genetic algorithm is used in this work to *estimate* the missing values by optimising an objective function as presented shortly in this section. The complete vector combining the estimated and the observed values is input into the autoencoder as shown in Figure 3. Symbols $X_k$ and $X_u$ represent the known variables and the unknown (or missing) variables respectively. The combination of $X_k$ and $X_u$ represent the full input space.

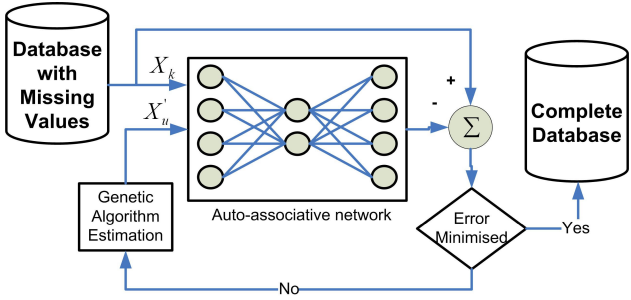

Figure 3: Autoencoder and GA Based missing data estimator structure

Considering that the method proposed here uses an autoencoder, one will expect the input to be very similar to the output for a well chosen architecture of the autoencoder. This is, however, only expected on a data set similar to the problem space from which the inter-correlations have been captured. The difference between the target and the actual output is used as the error and this error is defined as follows:

$$\varepsilon = \vec{x} - f(\vec{W}, \vec{x}) \qquad (3)$$

where $\vec{x}$ and $\vec{W}$ are input and weight vectors respectively. To make sure the error function is always positive, the square of the equation is used. This leads to the following equation:

$$\varepsilon = (\vec{x} - f(\vec{W}, \vec{x}))^2 \qquad (4)$$

Since the input vector consist of both the known, $X_k$ and unknown, $X_u$ entries, the error function can be written as follows:

$$\varepsilon = \left( \begin{Bmatrix} X_k \\ X_u \end{Bmatrix} - f\left( \begin{Bmatrix} X_k \\ X_u \end{Bmatrix}, w \right) \right)^2 \qquad (5)$$

and this equation is used as the objective function that is minimized using GA.

## 5. PROPOSED METHOD: ENSEMBLE BASED TECHNIQUE FOR MISSING DATA

The algorithm proposed here uses an ensemble of neural networks to perform both classification and regression in the presence of missing data. Ensemble based approaches have well been researched and have been found to improve classification performances in various applications [14-15]. The potential of using ensemble based approach for solving the missing data problem remains unexplored in both classification and regression problems. In the proposed method, batch training is performed whereas testing is done online. Training is achieved using a number of neural networks, each trained with a different combination of features. For a condition monitoring system that contains *n* sensors, the user has to state the value of $n_{avail}$, which is the number of features most likely to be available at any given time. Such information can be deduced from the reliability of the sensors as specified by manufacturers. Sensor manufacturers often state specifications such as *Mean-time-between failures* (MTBF) and *Mean-time-to-failure* (MTTF) which can help in detecting which sensors are most likely to fail than others. MTTF is used in cases where a sensor is replaced after a failure, whereas MTBF denotes time between failures where the sensor is repaired. There is nevertheless, no guarantee that failures will follow manufacturers' specifications.

When the number of sensors most likely to be available has been determined, the number of all possible networks can be calculated using:

$$N = \binom{n}{n_{avail}} = \frac{n!}{n(n-n_{avail})!} \qquad (6)$$

where $N$ is the total number of all possible networks, $n$ is the total number of features and $n_{avail}$ is the number of features most likely to be available at any time. Although the number $n_{avail}$ can be statistically calculated, it has an effect on the number of networks that can be available. Let us consider a simple example where the input space has 5 feature, labelled : *a*, *b*, *c*, *d* and *e* and there are 3 features that are most likely to be available at any time. Using equation (6), variable $N$ is found to be 10. These classifiers will be trained with features [*abc*, *abd*, *abe*, *acd*, *ace*, *ade*, *bcd*, *bce*, *bde*, *cde*]. In a case where one variable is missing, say, *a*, only four networks can be used for testing, and these are the classifiers that do not use *a* in their training input sequence. If we get a situation where two variables are missing, say *a* and *b*, we remain with one classifier. As a result, the number of classifiers reduces with an increase in a number of missing inputs per instance.

Each neural network is trained with $n_{avail}$ features. The validation process is then conducted and the outcome is used to decide on the combination scheme. The training process requires complete data to be available as training is done off-line. The available data set is divided into the 'training set' and the 'validation set'. Each network created is tested on the validation set and is assigned a weight according to its performance on the validation set. A diagrammatic illustration of the proposed ensemble approach is presented in Figure 4.

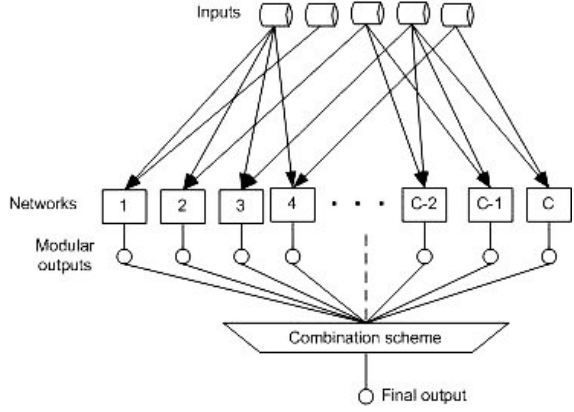

Figure 4: Diagrammatic illustration of the proposed ensemble based approach for missing data

For a classification task, the weight is assigned using the weighted majority scheme given by [16] as:

$$\alpha_i = \frac{1-E_i}{\sum_{j=1}^{N}(1-E_i)} \quad (7)$$

where $E_i$ is the estimate of model $i$'s error on the validation set. This kind of weight assignment has its roots in what is called boosting and is based on the fact that a set of networks that produces varying results can be combined to produce better results than each individual network in the ensemble [16]. The training algorithm is presented in *Algorithm 1* and the parameter $ntwk_i$ represents the $i^{th}$ neural network in the ensemble.

The testing procedure is different for classification and regression. In classification, testing begins by selecting an elite classifier. This is chosen to be the classifier with the best classification rate on the validation set. To this *elite* classifier, two more classifiers are gradually added, ensuring that an odd number is maintained. Weighted majority voting is used at each instance until the performance does not improve or until all classifiers are utilised. In a case of regression, all networks are used all at once and their predictions, together with their weights are used to compute the final value. The final predicted value is computed as follows:

$$f(x) = y \equiv \sum_{i=1}^{N} \alpha_i f_i(x) \quad (8)$$

where $\alpha$ is the weight assigned during the validation stage when no data were missing and $N$ is the total number of regressors. The parameter $\alpha$ is assigned such that $\sum_{i=1}^{N} \alpha_i = 1$. Considering that not all networks shall be available during testing, we define $N_{usable}$ as the number of regressors that are usable in obtaining the regression value of an instance $j$. As a result $\sum_{i=1}^{N_{usable}} \alpha_i \neq 1$.

We try to solve this by recalculating the weights such that the sum of all weights corresponding to $N_{usable}$ is 1.

**Algorithm 1**: Proposed algorithm for classification tasks

**input** : all variable $\in$ InputSpace & $n_{avail}$ obtained from the user
**output**: class
Calculate number of maximum Networks $N$ using Eq (6)
**forall** ($variables\ 1 \to X_n$) **do**
  Create all possible networks, $ntwk_1 \to ntwk_C$, each with $n_{avail}$ inputs
**end**
**while** *Training* **do**
  $\leftarrow$ Train $ntwk_i$ with a different combination of $n_{avl}$ inputs
  **forall** $i \to C$ **do**
    $\leftarrow$ Subject $ntwk_i$ to a validation set as follows:
    $\longrightarrow$ Select the corresponding features used;
    $\longrightarrow$ Obtain network performance;
    $\longrightarrow$ Assign weights, $\alpha$ according to Eq (7) and store for future use
  **end**
**end**
**while** *Testing* **do**
  $\leftarrow$ Load parameters from trainning;
  **if** *A Classification problem* **then**
    **foreach** *instance with missing values* **do**
      $\leftarrow$ Select networks, starting with those with bigger $\alpha$;
      $\leftarrow$ Bring 2 more networks, using their $\alpha$ as the selection criteria;
      $\leftarrow$ Use majority voting to obtain the final classification
    **end**
  **end**
  **if** *A Regression Problem* **then**
    **foreach** *instance with missing values* **do**
      $\leftarrow$ Get regression estimates from all networks trained without the current missing variable
      $\leftarrow$ Use their weights to compute the final value.
    **end**
  **end**
**end**

## 6. EXPERIMENTAL RESULTS AND DISCUSSION

This section presents the results obtained in the experiments conducted using the two techniques presented above. Firstly, the results of the proposed technique in a classification problem will be presented and later the method will be tested in a regression problem. In both cases, the results are compared to those obtained after imputing the missing values using the neural network-genetic algorithm combination as discussed above.

*6.1 Application to classification*

*Data set:* The experiment was performed using the Dissolved Gas Analysis (DGA) data obtained from a transformer bushing operating on-site. The data consist of 10 features, which are the gases that dissolved in the oil. The hypothesis in this experiment is to determine if the bushing condition (faulty or healthy) can be determined while some of the data are missing. The data was divided into the training set and the validation, each containing 2000 instances.

*Experimental setup:* The classification test was implemented using an ensemble of Fuzzy-ARTMAP networks. Two inputs were considered more likely to be missing and as a result, 8 were considered most likely to be available. The online process was simulated where data is sampled one instance at a time for testing. The network parameters were empirical determined and the vigilance parameter of 0.75 was used for the Fuzzy-ARTMAP. The results obtained were compared to those obtained using the the NN-GA approach, where for the GA, the crossover rate of 0.1 was used over 25 generations, each with a population size of 20. All these parameters were empirically determined.

*Results:* Using equation (6), a total of 45 networks was found to be the maximum possible. The performance was calculated only after 4000 cases have been evaluated and is shown in Figure 5. The classification increases with an increase in the number of classifiers used. Although all these classifiers were not trained with all the inputs, their combination seems to work better than one network. The classification accuracy obtained under missing data goes as high as 98.2% which compares very closely to a 100 % which is obtainable when no data is missing.

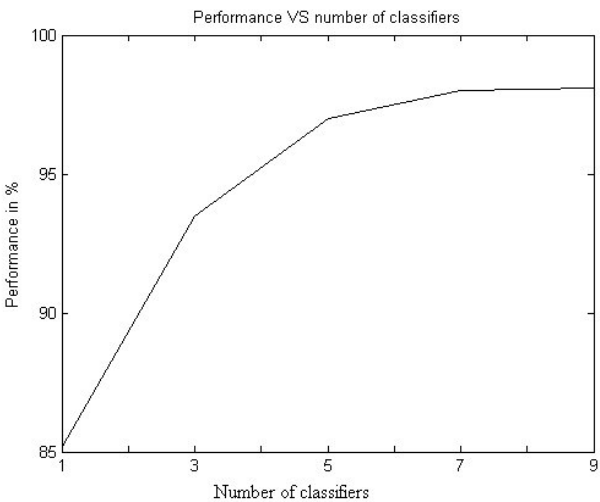

Figure 5: Diagrammatic illustration of the proposed ensemble based approach for missing data

Using the NN-GA approach, a classification of 96% was obtained. Results are tabulated in Table 1 below.

Table 1: Comparison between the proposed method and the NN-GA approach

|  | Proposed Algorithm | | NN-GA | |
|---|---|---|---|---|
| Number of missing | 1 | 2 | 1 | 2 |
| Accuracy (%) | 98.2 | 97.2 | 99 | 89.1 |
| Run time (s) | 0.86 | 0.77 | 0.67 | 1.33 |

The results presented in Table 1 clearly show that the proposed algorithms can be used as a means of solving the missing data problem. The proposed algorithm compares very well to the well know NN-GA approach. The run time for testing the performance of the method varies considerably. It can be noted from the table that for the NN-GA method, run time increase with increasing number of missing variables per instance. Contrary to the NN-GA, our proposed method offers run times that decrease with increasing number of inputs. The reason for this is that the number of Fuzzy-ARTMAP networks available reduces with an increasing number of inputs as mentioned earlier. However, this improvement in speed comes at a cost of the diversity. We tend to have less diversity as the number of training inputs increase. Furthermore, this method will completely come to a failure in a case where more than $n_{avl}$ inputs will be missing at the same time.

*6.1 Application to regression*

In this section, we extend the algorithm implemented in the above section to a regression problem. Instead of using an ensemble of Fuzzy ARTMAP networks as in classification, MLP networks are used. The reasons for this practice are two fold; firstly because MPL's are excellent regressors and secondly, to show that the proposed algorithm can be used with any architecture of neural networks.

*Database:* The data from a model of a Steam Generator at Abbott Power Plant [17] was used for this task. This data has four inputs, which are the *fuel, air, reference level* and the *disturbance* which is defined by the load level. There are two outputs which we shall try to predict using the proposed approach in the presence of missing data. These outputs are *drum pressure* and the *steam flow*.

*Experimental setup:* Although Fuzzy-ARTMAP could not be used for regression, we extended the same approach proposed above using MLP neural networks for regression problem. As before, this work regresses in order to obtain two outputs which are the *drum pressure* and the *steam flow*. We assume $n_{avl} = 2$ is the case and as a result, only two inputs can be used. We create an ensemble of MLP networks, each with five hidden nodes and trained only using two of the inputs to obtain the output. Due to limited features in the data set, this work shall only consider a maximum of one sensor failure per instance. Each network was trained with 1200 training cycles using the scaled conjugate gradient algorithm and a hyperbolic tangent activation function. All these training parameters were again empirically determined.

*Results:* Since testing is done online where one input arrives at a time, evaluation of performance at each instance would not give a general view of how the algorithm works. The work therefore evaluates the general performance using the following formula only after *N* instances have been predicted.

$$Error = \frac{n_\tau}{N} \times 100\% \qquad (9)$$

where $n_\tau$ is the number of predictions within a certain tolerance. In this paper, a tolerance of 20% is used and was arbitrarily chosen. Results are summarized in Table 2

Table 2: Regression accuracy obtained without estimating the missing values.

|  | Proposed Algorithm | | NN-GA | |
|---|---|---|---|---|
| Number of missing | 1 | 2 | 1 | 2 |
|  | Perf (%) | Time | Perf (%) | Time |
| Drum Pressure | 98.2 | 97.2 | 68 | 126 |
| Steam Flow | 86 | 0.77 | 84 | 98 |

'Perf' indicates the accuracy in percentage whereas time indicates the running time in seconds. Results show that the proposed method is well suited for the problem under investigation. The proposed method performs better than the combination of GA and autoencoder neural networks in the regression problem under investigation. The reason is that the errors that are made when inputting the missing data in the NN-GA approach are further propagated to the output-prediction stage. The ensemble based approach proposed here does not suffer from this problem as there is no attempt to approximate the missing variables. It can also be observed that the ensemble based approach takes less time that the NN-GA method. The reason for this is that GA may take longer times to converge to reliable estimates of the missing values depending on the objective function to be optimised. Although, the prediction times are negligibly small, an ensemble based technique takes more time to train since training involves a lot of networks.

## 7. CONCLUSION

In this paper a new techniques for dealing with missing data for online condition monitoring problem was presented and studied. Firstly the problem of classifying in the presence of missing data was addressed, where no attempts are made to recover the missing values. The problem domain was then extended to regression. The proposed technique performs better than the NN-GA approach, both in accuracy and time efficiency during testing. The advantage of the proposed technique is that it eliminates the need for finding the best estimate of the data, and hence, saves time. Future work will explore the incremental learning ability of the Fuzzy ARTMAP in the proposed algorithm.

**Acknowledgements**
The financial assistance of the National Research Foundation (NRF) of South Africa and the Carl and Emily Fuchs Foundation is hereby acknowledged.